\documentclass[11pt]{article}

\usepackage[preprint]{acl}

\usepackage{times}
\usepackage{latexsym}
\usepackage[T1]{fontenc}
\usepackage[utf8]{inputenc}
\usepackage{microtype}
\usepackage{inconsolata}
\usepackage{graphicx}

\usepackage{amsmath}
\usepackage{amssymb}
\usepackage{amsfonts}
\usepackage{booktabs}
\usepackage{colortbl}
\usepackage{multirow}
\usepackage{xspace}

\title{TempoMoE: Identifying and Steering Fast--Slow Thinking
Experts for Efficient Multimodal Reasoning}

\author{
  Xiyue Zhang\textsuperscript{1}\quad
  Keru Zhou\textsuperscript{2}\quad
  Ziyi Wang\textsuperscript{1}\quad
  Bing Hu\textsuperscript{1}\quad
  Li Wang\textsuperscript{3} \\
  \textsuperscript{1}Zhejiang University\quad
  \textsuperscript{2}Tsinghua University\quad
  \textsuperscript{3}Wenzhou Business College
}

\begin{document}

\maketitle
\raggedbottom

\begin{abstract}

Multimodal Mixture-of-Experts (MoE) reasoning models rely on long
chain-of-thought (CoT) for complex tasks, yet often overthink easy
inputs, increasing token cost and latency, and sometimes degrading
accuracy.
Existing methods optimize reasoning length during training or impose
external control at inference, but leave the underlying reasoning
mechanism a black box.
We find that fast and slow thinking exhibit distinct expert-routing
patterns: holding the input fixed and varying only the prompted
reasoning tempo produces stable routing shifts.
Suppressing slow-thinking experts shortens the reasoning chain,
whereas suppressing fast-thinking experts lengthens it, providing
causal evidence that these experts influence reasoning depth.
Based on this finding, we propose \textbf{TempoMoE}, a training-free,
two-stage framework.
It identifies fast- and slow-thinking experts via tempo-contrastive
routing analysis and redistributes routing weights within the original
top-$k$ set according to sample difficulty.
It amplifies fast-thinking experts and suppresses slow-thinking
experts on easy inputs and reverses this adjustment on hard ones,
enabling difficulty-adaptive control over reasoning depth.
We evaluate TempoMoE on Kimi-VL, Qwen3-VL, and ERNIE-4.5-VL across
six benchmarks.
Compared with Vanilla, TempoMoE consistently improves accuracy while
reducing output tokens, with accuracy gains of up to 2.6 points and
token reductions of up to 35.4\%.

\end{abstract}

\section{Introduction}
\label{sec:introduction}

\begin{figure}[t]
    \centering
    \includegraphics[width=\columnwidth]{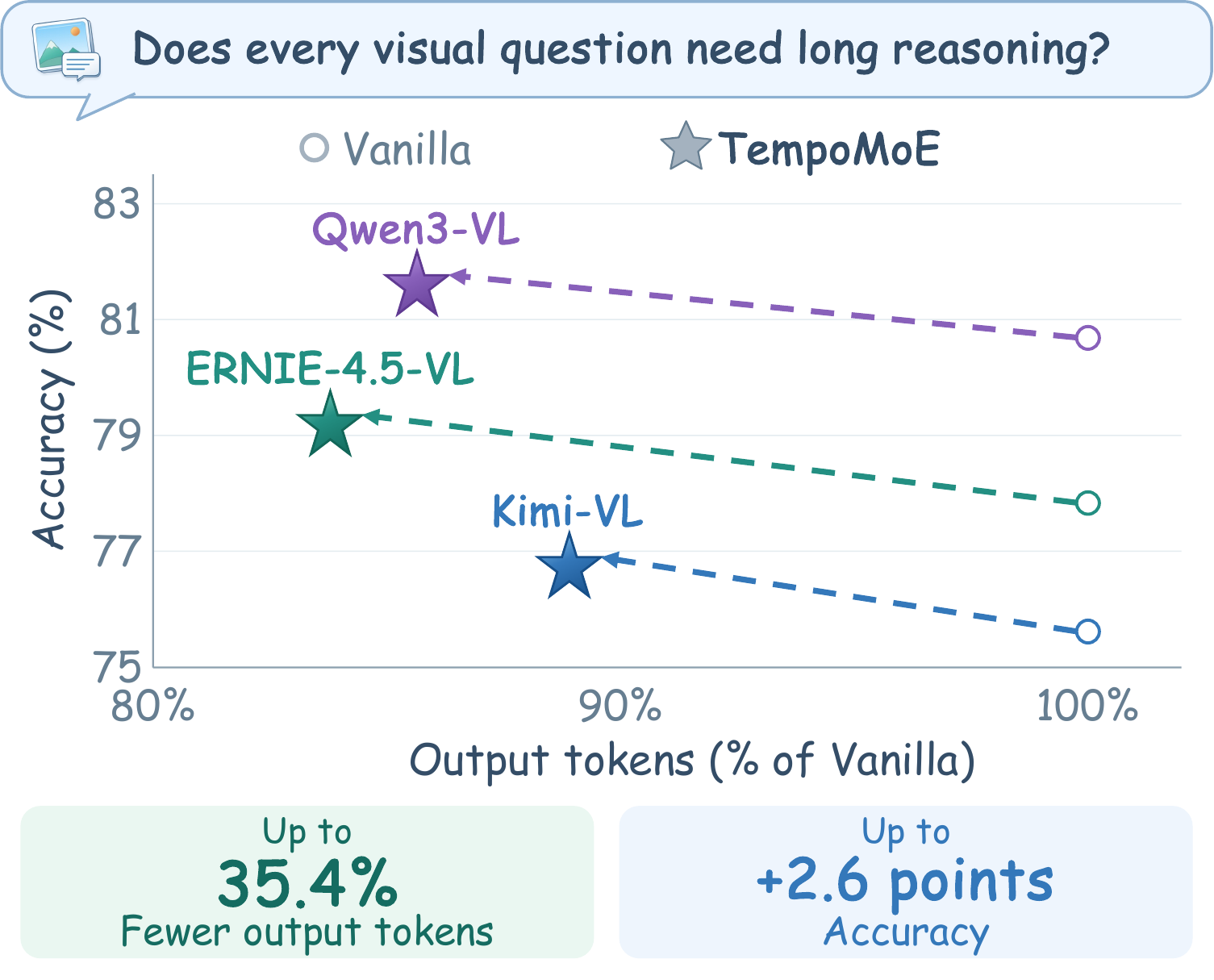}
    \caption{Average accuracy and output tokens over six benchmarks.
    Tokens are relative to Vanilla; bottom values show maximum gains.}
    \label{fig:teaser}
\end{figure}

Large reasoning models rely on chain-of-thought (CoT)
reasoning to solve complex tasks~\cite{wei2022chain}.
Yet the required reasoning effort varies substantially across inputs.
In multimodal tasks, object recognition and multi-step visual
mathematics require different levels of computation.
However, models may repeatedly verify simple perceptual judgments,
increasing token cost and latency and potentially degrading answer
quality through redundant reasoning and error
accumulation~\cite{chen2024overthinking,sui2025stopoverthinking}.
Matching computation to input demands is therefore important
for efficient multimodal reasoning.

Existing work mitigates overthinking by optimizing reasoning length
during training or controlling generation at inference through
prompting, token budgets, or decoding
heuristics~\cite{huang2024harder,xiao2025fastslow,diao2026gpro,
huang2026cgrs,yang2025deer}.
The MoE architecture provides a natural entry point for controlling
internal computational pathways.
Its sparse router dispatches each token to a small subset of
experts~\cite{shazeer2017sparsely,fedus2022switch},
making expert selection and contributions observable and steerable.
Prior work has demonstrated expert specialization and the value of
targeted expert
interventions~\cite{wang2025rice,fayyaz2025steering,hu2026activated}.
However, routing differences between fast and slow thinking
for a fixed image and question, and their use for adaptive
reasoning control, remain underexplored.

We first conduct a controlled comparison, holding the image and
question fixed while varying only the prompted reasoning tempo.
We observe stable routing differences during question processing
and decoding.
Targeted interventions further show that masking the identified
slow experts shortens outputs, whereas masking fast experts
lengthens them.
These findings link expert routing to generation behavior
and motivate inference-time expert steering.

Based on these findings, we propose \textbf{TempoMoE}, a training-free,
two-stage framework.
In the first stage, we identify fast-slow experts through routing
comparisons, robustness screening, and intervention tests.
In the second stage, TempoMoE estimates input difficulty using
prefill routing entropy and continuously adjusts weights within
the original top-$k$ expert set.
It favors fast experts on easy inputs and slow experts on hard inputs,
without changing the selected expert set or updating model parameters.

We evaluate TempoMoE on three multimodal MoE reasoning models
(Kimi-VL, Qwen3-VL, and ERNIE-4.5-VL) across six benchmarks,
comparing it with Vanilla and four training-free baselines.
Figure~\ref{fig:teaser} summarizes average accuracy and token usage.
Compared with Vanilla, TempoMoE improves accuracy while reducing
output tokens in all 18 model--benchmark combinations,
with accuracy gains of up to 2.6 points and token reductions
of up to 35.4\%.

Our contributions are threefold:
\begin{itemize}
    \item To our knowledge, we are the first to uncover distinct
    fast- and slow-thinking experts in multimodal MoE reasoning
    models and provide intervention-based evidence that they
    influence reasoning depth through role-consistent interventions.

    \item We propose TempoMoE, a training-free, two-stage framework
    that first identifies fast- and slow-thinking experts by
    contrasting their routing patterns across reasoning tempos,
    and then adaptively redistributes their routing weights within
    the original top-$k$ set according to sample difficulty.

    \item We validate TempoMoE across three multimodal MoE reasoning
    models with distinct routing designs and six benchmarks,
    showing that it consistently improves accuracy while reducing
    output tokens.
\end{itemize}

\section{Related Work}
\label{sec:related}

\paragraph{Efficient reasoning and overthinking mitigation.}
Long chain-of-thought can improve performance on complex tasks but often
introduces redundant computation on easy inputs. Existing approaches broadly
fall into training-based reasoning-budget optimization and inference-time
control. Training-based methods learn concise or sample-adaptive reasoning
through length optimization, reinforcement learning, adaptive CoT triggering,
or budget prediction~\cite{arora2025efficient,aggarwal2025l1,lou2025adacot,
wu2025arm,li2025selfbudgeter}. Inference-time methods reduce redundant
generation through concise intermediate reasoning~\cite{xu2025chainofdraft},
prompt-based token budgets or budget forcing~\cite{han2025tale,
muennighoff2025s1}, confidence-based early exit~\cite{yang2025deer,
fu2024dynasor}, or suppression of reflection triggers~\cite{huang2026cgrs}.
These methods highlight the value of adapting computation to input difficulty
but primarily control observable generation behaviors, including reasoning
budgets, output length, and stopping time, without identifying the internal
pathways that produce them.

\paragraph{Fast--slow reasoning in multimodal models.}
Fast--slow reasoning has received less attention in multimodal models than in
the text-only setting. FAST-GRPO~\cite{xiao2025fastslow},
DualMindVLM~\cite{lin2025dualmindvlm}, and GPRO~\cite{diao2026gpro} adapt the
reasoning effort of vision--language models through difficulty-aware training,
reasoning-mode selection, or branch routing. Metis-HOME~\cite{lan2025metishome}
further incorporates fast--slow reasoning into a trained multimodal MoE
architecture. These methods typically rely on retraining or architectural
changes and control reasoning through sample-level decisions, discrete modes,
or separate branches. They do not explicitly identify whether fast and slow
reasoning correspond to distinct internal components.

\paragraph{Expert specialization and routing intervention in MoE.}
Mixture-of-Experts (MoE) architectures provide a natural setting for studying
internal mechanisms because sparse routing exposes expert-level computation.
They have been increasingly adopted in vision--language models to scale model
capacity with limited active computation~\cite{lin2024moellava,li2025unimoe,
wu2024deepseekvl2}. Recent studies analyze routing patterns to characterize
expert specialization across tasks, domains, languages, semantic content,
safety properties, and cognitive effort~\cite{lo2025closer,
olson2025semantic,hu2026activated,bandarkar2025multilingual,lai2025safex,
wang2025rice}. Other approaches manipulate MoE behavior through expert
fine-tuning, activation or deactivation, test-time rerouting, and dynamic
expert skipping~\cite{wang2024esft,fayyaz2025steering,li2025r2t2,
xu2026seeing,huang2025modes}. Fast--slow reasoning presents a different setting
from task or domain specialization with explicit labels: the same model can
exhibit distinct reasoning states for the same image and question. TempoMoE
identifies experts associated with these states through controlled
reasoning-tempo comparisons, validates their functional roles through targeted
interventions, and uses them for difficulty-adaptive reasoning control.

\begin{figure*}[t]
\centering
\includegraphics[width=\textwidth]{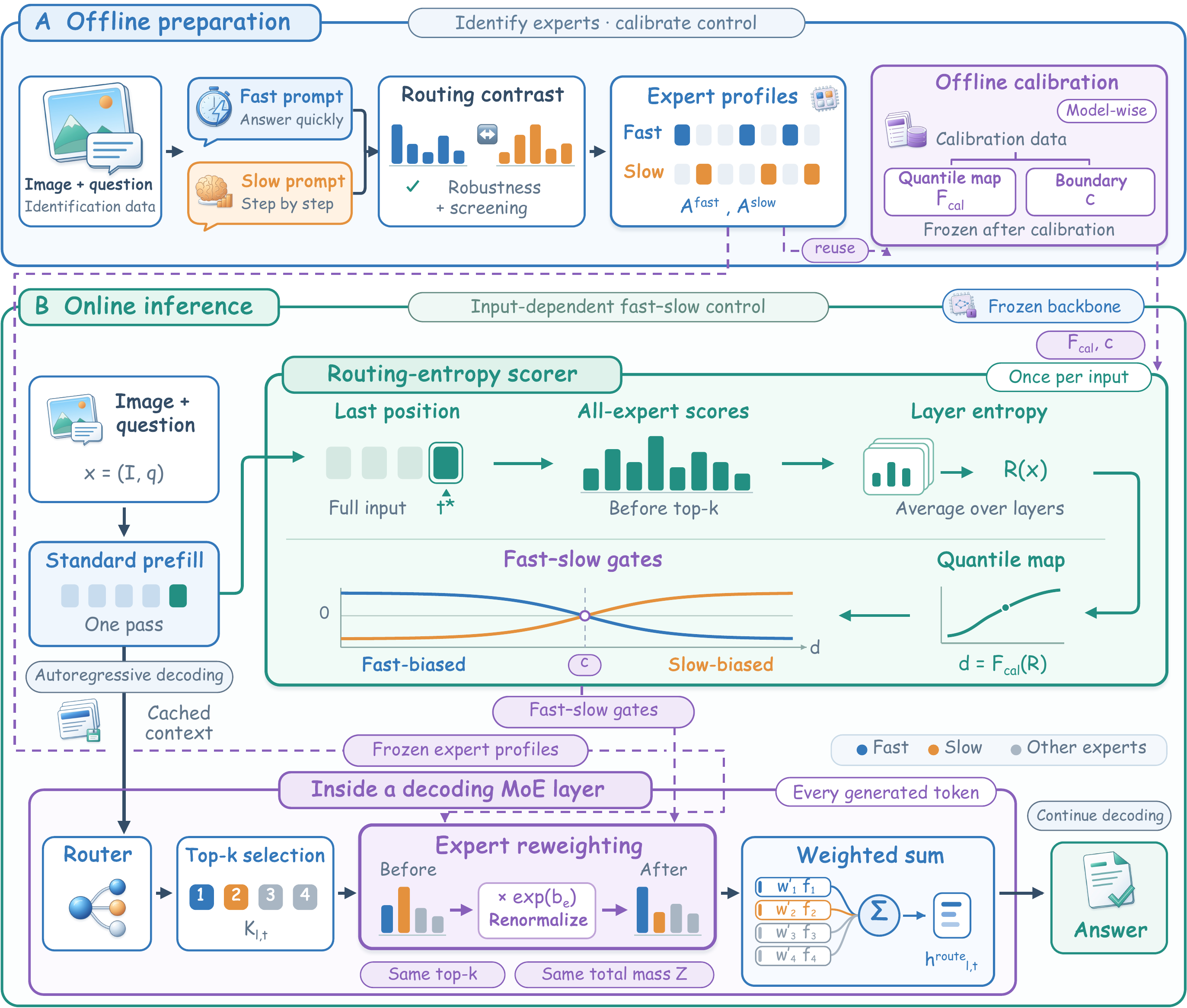}
\caption{Overview of TempoMoE.
(A) Fast-slow experts are identified offline by contrasting routing patterns
under paired prompts.
(B) Prefill routing entropy determines fast-slow gates that reweight the
original top-$k$ experts during decoding, preserving the selected expert set,
total routing mass, and all model parameters.}
\label{fig:pipeline}
\end{figure*}

\section{Identifying Fast-Slow Experts}
\label{sec:method}

TempoMoE consists of two stages: offline expert identification and adaptive
modulation during inference, as illustrated in Figure~\ref{fig:pipeline}.
In Stage~1, we identify fast-slow experts across MoE layers by contrasting routing
patterns under paired prompts. In Stage~2, TempoMoE estimates input difficulty
and performs soft reweighting within the original top-$k$ expert set.
This enables continuous modulation of reasoning effort without updating model
parameters or changing the router's expert selection.

\subsection{MoE Routing and Intervention Design}
\label{sec:method_prelim}

\paragraph{MoE routing.}
For token $t$ at layer $l$, the sparse router selects a top-$k$ expert set
$\mathcal{K}_{l,t}$ and assigns a routing weight $w_{l,t,e}$ to each selected
expert. The output of the routed expert branch is
\begin{equation}
h^{\mathrm{route}}_{l,t}
=
\sum_{e\in\mathcal{K}_{l,t}}
w_{l,t,e}\cdot E_{l,e}\!\left(h^{\mathrm{in}}_{l,t}\right).
\label{eq:moe_layer}
\end{equation}
Shared experts, when present, remain unchanged and are excluded from both
expert identification and routing modulation.

\paragraph{Routing intervention design.}
We distinguish two intervention levels. In Stage~1, a
\emph{selection-level intervention} masks candidates before top-$k$ selection
to assess their effects on output length. In Stage~2, a
\emph{weight-level intervention} redistributes routing weights after top-$k$
selection, preserving the original expert set $\mathcal{K}_{l,t}$.

\subsection{Fast-Slow Expert Identification}
\label{sec:method_identification}

\paragraph{Routing collection.}
To identify fast-slow experts, we keep the image and question fixed and use
different prompts to elicit fast or slow thinking. Given a multimodal sample
$x=(I,q)$, we construct fast and slow prompt sets $\mathcal{P}_{f}$ and
$\mathcal{P}_{s}$ with the same answering objective. We then record the
selected experts and routing weights at each MoE layer.

We use three fast prompts requesting direct answers without explicit reasoning
and three slow prompts requesting step-by-step reasoning. For example, one pair
begins with ``Look at the image and answer the question directly'' and
``Carefully inspect the image and solve the question step by step,''
respectively. All prompts specify the same final-answer format:
\texttt{Final answer: \textless answer\textgreater}.

We divide the input and generated sequence into image, question, instruction,
and decode segments. The instruction segment contains the tempo directive and
is used only for diagnostic analysis. The image segment serves as a control
because the visual input is identical under both conditions. Candidate
identification uses routing differences in the question and decode segments.

\paragraph{Routing differences.}
Let $w_{l,t,e}(x,p)$ denote the routing weight of expert $e$ at layer $l$ for
token $t$ under prompt $p$. Its effective routing contribution is
\begin{equation}
\gamma_{l,e}(t;x,p)
=
w_{l,t,e}(x,p)\cdot
\mathbf{1}\!\left[e\in\mathcal{K}_{l,t}(x,p)\right].
\label{eq:effective_contribution}
\end{equation}
For segment $S$, we average this contribution over its tokens, the
identification set $\mathcal{D}$, and prompt set $\mathcal{P}$:
\begin{equation}
\begin{aligned}
M^{S}_{l,e}(\mathcal{P})
&=\frac{1}{|\mathcal{D}||\mathcal{P}|}
\sum_{x\in\mathcal{D}}\sum_{p\in\mathcal{P}} \\
&\quad\frac{1}{|\mathcal{T}_{S}(x,p)|}
\sum_{t\in\mathcal{T}_{S}(x,p)}\gamma_{l,e}(t;x,p),
\end{aligned}
\label{eq:routing_mass_stage1}
\end{equation}
where $\mathcal{T}_{S}(x,p)$ denotes the tokens in segment $S$.
Length normalization prevents longer slow-thinking outputs from receiving
disproportionate weight. The routing difference is
\begin{equation}
\Delta M^{S}_{l,e}
=M^{S}_{l,e}(\mathcal{P}_{s})-M^{S}_{l,e}(\mathcal{P}_{f}).
\label{eq:delta_m}
\end{equation}
Positive and negative values indicate preferences for slow and fast prompts,
respectively.

\paragraph{Candidate selection.}
Let $y\in\{+1,-1\}$ denote slow and fast thinking, respectively.
Bootstrap stability $\pi^{S,y}_{l,e}$ measures how often an expert appears
among the leading candidates in direction $y$ across resampled datasets.
Prompt consistency $\rho^{S,y}_{l,e}$ measures the proportion of fast-slow
prompt pairs that preserve this directional preference. We filter out experts
with small routing differences or negligible average contributions and rank
the remaining candidates by
\begin{equation}
\mathrm{Score}^{S,y}_{l,e}
=\left(y\cdot\Delta M^{S}_{l,e}\right)
\cdot\pi^{S,y}_{l,e}\cdot\rho^{S,y}_{l,e}.
\label{eq:expert_score}
\end{equation}

\paragraph{Intervention screening.}
On the identification set, we mask each candidate before top-$k$ selection and
measure the resulting change in output length, including reasoning and
final-answer tokens. A slow candidate is retained if masking it shortens the
output, whereas a fast candidate is retained if masking it lengthens the
output. 

The retained candidates form the fast and slow expert sets
$\mathcal{E}^{\mathrm{fast}}$ and $\mathcal{E}^{\mathrm{slow}}$.
We normalize their ranking scores within each layer to $[0,1]$, yielding
$A^{\mathrm{fast}}_{l,e}$ and $A^{\mathrm{slow}}_{l,e}$ for Stage~2.
All other experts receive a profile score of zero. The expert sets and profiles
are fixed before held-out validation and downstream evaluation.

\section{Difficulty-Adaptive Expert Steering}
\label{sec:method_modulation}

Given the fast-slow expert profiles identified in Stage~1, TempoMoE
estimates input difficulty before decoding and uses it to continuously
modulate routing weights within the original top-$k$ expert set.

\paragraph{Sample difficulty estimation.}
TempoMoE estimates input difficulty using routing entropy from the standard
prefill pass. We collect routing scores at the last non-padding position
of the complete input, which can attend to the full image and question.
At each MoE layer, we normalize the native nonnegative routing scores
before top-$k$ selection to obtain a distribution $p_{l,e}(x)$ over all
routed experts. Shared experts and correction biases used solely for
expert selection are excluded. We then compute normalized entropy and
average it across layers:
\begin{equation}
R(x)
=
-\frac{1}{|\mathcal{L}|}
\sum_{l\in\mathcal{L}}
\frac{1}{\log E_l}
\sum_{e=1}^{E_l}
p_{l,e}(x)\log p_{l,e}(x),
\label{eq:routing_entropy}
\end{equation}
where $\mathcal{L}$ denotes the set of MoE layers with at least two routed
experts, and $E_l$ is the number of routed experts at layer $l$.
Larger values of $R(x)$ indicate more dispersed routing distributions.

For each model, we construct an empirical percentile mapping
$F_{\mathrm{cal}}$ from routing entropy on its calibration set, yielding
the difficulty score $d(x)=F_{\mathrm{cal}}(R(x))\in[0,1]$.
The mapping remains fixed throughout validation and testing.
This score serves as a proxy for reasoning effort, not a probability
of answer correctness.

\paragraph{Adaptive fast-slow modulation.}
We map the difficulty score to a continuous control signal centered at
a boundary $c$:
\begin{equation}
s(x)
=
\tanh\!\left(\frac{d(x)-c}{\tau}\right).
\label{eq:difficulty_control}
\end{equation}
The modulation coefficients for fast and slow experts are
\begin{equation}
g_{\mathrm{fast}}(x)
=
-g_{\max}s(x),
\qquad
g_{\mathrm{slow}}(x)
=
g_{\max}s(x).
\label{eq:symmetric_gate}
\end{equation}
Here, $c$ is the boundary between fast and slow modulation.
Across all models we use $c=0.5$, $\tau=0.25$, and $g_{\max}=1.0$, select the top two experts per direction for each model, and apply reweighting only during
decoding. The parameter $\tau>0$ controls the transition sharpness around
the boundary, while $g_{\max}\geq0$ determines the maximum modulation strength.

When $d(x)<c$, TempoMoE amplifies fast experts and suppresses slow experts;
the direction is reversed when $d(x)>c$. The modulation weakens near
the boundary, and both coefficients vanish at $d(x)=c$, leaving the
original routing weights unchanged. Thus, the modulation varies
continuously with input difficulty rather than switching between
two discrete modes.

\paragraph{Soft reweighting within the top-$k$ set.}
We combine the fast-slow gates with the expert profiles to obtain a bias
for expert $e$ at layer $l$ for each input:
\begin{equation}
b_{l,e}(x)
=
g_{\mathrm{fast}}(x)A^{\mathrm{fast}}_{l,e}
+
g_{\mathrm{slow}}(x)A^{\mathrm{slow}}_{l,e}.
\label{eq:expert_bias}
\end{equation}

For the expert set $\mathcal{K}_{l,t}$ selected by the original router,
TempoMoE applies this bias and renormalizes the weights within the set:
\begin{equation}
w'_{l,t,e}
=
Z_{l,t}
\frac{
w_{l,t,e}\exp\!\left(b_{l,e}(x)\right)
}{
\sum_{j\in\mathcal{K}_{l,t}}
w_{l,t,j}\exp\!\left(b_{l,j}(x)\right)
},
\qquad
e\in\mathcal{K}_{l,t},
\label{eq:topk_reweighting}
\end{equation}
where $w_{l,t,e}$ is the original routing weight and
$Z_{l,t}=\sum_{j\in\mathcal{K}_{l,t}}w_{l,t,j}$ is the total routing mass
before reweighting. This operation changes only the relative contributions
of the selected experts, preserving both the expert set and its total
routing mass. Experts not identified as fast or slow receive zero bias,
and shared experts remain unchanged.

\paragraph{Online computational cost.}
The difficulty score is computed once per input during the standard
prefill pass and remains fixed throughout decoding. This requires neither
trial generation nor an additional model forward pass. Normalization and
entropy calculation use FP32 and add
$O(\sum_{l\in\mathcal{L}}E_l)$ operations, plus one percentile mapping lookup.
Reweighting is applied only during decoding. It adds $O(k)$ scalar
operations per routed MoE layer and generated token, with no additional
expert feed-forward computation.

\section{Experiments}
\label{sec:experiments}

Experiments span three multimodal MoE reasoning models with distinct
routing designs and six benchmarks covering a range of reasoning demands.
We examine TempoMoE's trade-off between accuracy and efficiency across models
and tasks, with ablations isolating the contributions of expert identification,
difficulty-adaptive modulation, and top-$k$-preserving soft reweighting.

\subsection{Experimental Setup}
\label{sec:exp_setup}

\paragraph{Models.}
TempoMoE is evaluated on three multimodal MoE reasoning models:
Kimi-VL-A3B-Thinking-2506~\cite{kimiteam2025kimivl},
Qwen3-VL-30B-A3B-Thinking~\cite{bai2025qwen3vl}, and
ERNIE-4.5-VL-28B-A3B-Thinking~\cite{ernie2025technicalreport}.
Their differences in expert count, routing sparsity, and shared-expert design
allow us to assess generalization across distinct MoE routing architectures.
All experiments use the corresponding Thinking checkpoints with BF16
inference, while shared generation settings remain fixed across methods.

\paragraph{Benchmarks.}
The benchmark suite comprises six multimodal tasks spanning direct visual
understanding to multi-step reasoning. OCRBench~\cite{liu2023ocrbench},
RealWorldQA~\cite{xai2024realworldqa}, and
ChartQA~\cite{masry2022chartqa} emphasize visual recognition and
information extraction; MMStar~\cite{chen2024mmstar} jointly evaluates
multimodal perception and reasoning; MathVista~\cite{lu2024mathvista}
targets visual mathematical reasoning, whereas
MMMU~\cite{yue2024mmmu} evaluates multidisciplinary knowledge and reasoning.

\paragraph{Baselines.}
We compare TempoMoE with Vanilla and four representative training-free
baselines. NoThinking~\citep{ma2025nothinking} skips explicit thinking,
TALE~\citep{han2025tale} allocates an input-dependent reasoning budget,
and NoWait~\citep{wang2025nowait} suppresses redundant reflection during
decoding. RICE~\citep{wang2025rice} identifies cognitive experts through
associations between expert activations and thinking-related tokens.
It amplifies their routing weights at inference time, providing a baseline for expert-level reasoning control.

\paragraph{Metrics.}
We report accuracy (Acc, \%) and average output tokens (Tok), including both
the reasoning segment and the final answer. Tok serves as a proxy for
autoregressive generation cost. For difficulty-stratified analysis, we divide
evaluation samples into Easy and Hard using the calibration-set median of
$d(x)$, without evaluation labels. This split is used only for analysis;
online modulation still follows the continuous control signal centered at
$0.5$ in Equation~\eqref{eq:difficulty_control}.

\subsection{Main Results}
\label{sec:exp_main}

\paragraph{Expert identification.}
We use 600 samples (100 per benchmark) for expert identification and
difficulty calibration. Each sample is run with three fast- and three
slow-thinking prompts, yielding 3{,}600 routing runs per model. A separate
600-sample set, disjoint from both the identification and main evaluation
sets, is used for intervention validation under the standard Thinking prompt
with Vanilla, fast-expert masking, and slow-expert masking.
Table~\ref{tab:identified_experts} reports the identification and
intervention results.

\begin{table}[!htbp]
\centering
{\footnotesize
\setlength{\tabcolsep}{1.35pt}
\renewcommand{\arraystretch}{1.02}
\begin{tabular}{@{}llrccr@{}}
\toprule
Model & Experts
& $\Delta M$
& Stab.
& Cons.
& $\Delta$Tok \\
\midrule
\multirow{2}{*}{Kimi}
& F: (8,6), (12,46)
& $-0.037$ & 100 & 94--100 & $+37.1$ \\
& S: (1,13), (8,37)
& $+0.041$ & 100 & 100 & $-51.7$ \\

\addlinespace[1.5pt]
\multirow{2}{*}{Qwen}
& F: (28,63), (33,49)
& $-0.028$ & 94 & 100 & $+32.1$ \\
& S: (28,79), (40,79)
& $+0.023$ & 100 & 100 & $-45.1$ \\

\addlinespace[1.5pt]
\multirow{2}{*}{ERNIE}
& F: (9,8), (8,57)
& $-0.027$ & 100 & 100 & $+24.0$ \\
& S: (18,19), (21,34)
& $+0.024$ & 98--100 & 94--100 & $-30.9$ \\
\bottomrule
\end{tabular}
}
\caption{Identified fast/slow experts and held-out validation.
F/S denote fast/slow experts; each pair denotes (layer, expert);
$\Delta M$ is averaged over the two experts; Stab./Cons. are percentages;
$\Delta$Tok is the mean token change after joint masking.}
\label{tab:identified_experts}
\end{table}

Across all three models, masking fast-thinking experts increases output
length, whereas masking slow-thinking experts shortens it. High bootstrap
stability and prompt consistency further support robust expert identification
and role-consistent intervention effects.

\paragraph{Performance across models.}
Table~\ref{tab:main} reports complete results for TempoMoE, Vanilla, and
the four training-free baselines. Compared with Vanilla, TempoMoE improves
accuracy while reducing output tokens in all 18 model--benchmark
combinations. When macro-averaged over the six benchmarks,
it outperforms Vanilla on both metrics for all three models. At the
benchmark level, accuracy improves by up to 2.6 points and output tokens
decrease by up to 35.4\%.

\begin{table*}[!t]
\centering
{\small
\setlength{\tabcolsep}{2.5pt}
\begin{tabular}{l cc cc cc cc cc cc}
\toprule
\multirow{2}{*}{Method}
& \multicolumn{2}{c}{MMStar}
& \multicolumn{2}{c}{OCRBench}
& \multicolumn{2}{c}{RealWorldQA}
& \multicolumn{2}{c}{MathVista}
& \multicolumn{2}{c}{ChartQA}
& \multicolumn{2}{c}{MMMU} \\
\cmidrule(lr){2-3}
\cmidrule(lr){4-5}
\cmidrule(lr){6-7}
\cmidrule(lr){8-9}
\cmidrule(lr){10-11}
\cmidrule(lr){12-13}
& Acc$\uparrow$ & Tok$\downarrow$
& Acc$\uparrow$ & Tok$\downarrow$
& Acc$\uparrow$ & Tok$\downarrow$
& Acc$\uparrow$ & Tok$\downarrow$
& Acc$\uparrow$ & Tok$\downarrow$
& Acc$\uparrow$ & Tok$\downarrow$ \\
\midrule

\rowcolor{gray!12}
\multicolumn{13}{c}{\textit{Kimi-VL-A3B-Thinking-2506}} \\
Vanilla
& 70.4 & 502.6
& 83.9 & 136.4
& 70.0 & 138.5
& 80.1 & 906.4
& 85.3 & 134.4
& 64.0 & 1934.8 \\
NoThinking
& 66.1 & 174.8
& 81.7 & 60.1
& 66.8 & 3.4
& 74.2 & 670.2
& 84.5 & 36.1
& 56.4 & 398.4 \\
TALE
& 65.6 & 457.5
& 74.6 & 119.6
& 67.8 & 127.2
& 74.9 & 836.3
& 80.9 & 122.0
& 60.8 & 1138.2 \\
NoWait
& 67.7 & 456.1
& \textbf{85.9} & 98.5
& 70.0 & 89.9
& 75.3 & 700.3
& 85.0 & 99.6
& 62.2 & 1246.3 \\
RICE
& 69.2 & 474.2
& 80.0 & 125.3
& 68.5 & 95.1
& 78.8 & 837.9
& 85.4 & 158.5
& 64.0 & 1835.2 \\
\textbf{TempoMoE}
& \textbf{71.8} & 466.6
& 85.3 & 107.2
& \textbf{71.9} & 111.7
& \textbf{81.0} & 747.3
& \textbf{86.1} & 122.4
& \textbf{64.5} & 1781.6 \\

\midrule
\rowcolor{gray!12}
\multicolumn{13}{c}{\textit{Qwen3-VL-30B-A3B-Thinking}} \\
Vanilla
& 75.5 & 782.8
& 83.9 & 168.6
& 77.4 & 246.5
& 81.9 & 1350.7
& 89.4 & 181.6
& 76.0 & 1556.7 \\
NoThinking
& 64.5 & 188.2
& 82.4 & 103.6
& 71.6 & 3.0
& 64.8 & 646.1
& 81.5 & 121.5
& 63.0 & 599.7 \\
TALE
& 73.3 & 644.9
& 84.2 & 137.6
& 73.9 & 217.7
& 72.1 & 782.5
& 87.3 & 137.9
& 72.1 & 1049.5 \\
NoWait
& 73.3 & 543.1
& 83.7 & 122.3
& 73.6 & 144.3
& 64.8 & 877.9
& 87.1 & 133.9
& 71.3 & 1211.3 \\
RICE
& 75.8 & 651.3
& 84.0 & 112.4
& 75.3 & 222.1
& 76.5 & 1137.4
& 84.0 & 165.8
& 71.0 & 1425.9 \\
\textbf{TempoMoE}
& \textbf{76.2} & 729.4
& \textbf{85.5} & 130.0
& \textbf{78.2} & 236.8
& \textbf{82.9} & 1172.9
& \textbf{90.3} & 149.2
& \textbf{76.7} & 1253.4 \\

\midrule
\rowcolor{gray!12}
\multicolumn{13}{c}{\textit{ERNIE-4.5-VL-28B-A3B-Thinking}} \\
Vanilla
& 68.9 & 858.3
& 86.4 & 118.2
& 69.9 & 332.2
& 82.5 & 1576.1
& 87.1 & 156.0
& 72.2 & 2081.1 \\
NoThinking
& 62.8 & 561.2
& 84.1 & 37.9
& 66.4 & 21.9
& 75.6 & 507.2
& 83.9 & 38.2
& 63.0 & 671.3 \\
TALE
& 69.0 & 636.4
& 86.5 & 99.6
& 69.5 & 293.0
& 80.1 & 994.7
& 86.2 & 38.1
& 68.4 & 1320.6 \\
NoWait
& 66.0 & 576.7
& 86.1 & 81.6
& 67.1 & 290.3
& 72.8 & 1218.2
& \textbf{89.6} & 125.6
& 65.7 & 1459.0 \\
RICE
& 69.3 & 693.0
& 87.3 & 142.7
& 66.0 & 187.1
& 78.2 & 1347.5
& 87.0 & 151.4
& 69.4 & 1729.8 \\
\textbf{TempoMoE}
& \textbf{69.7} & 671.8
& \textbf{89.0} & 76.4
& \textbf{71.3} & 263.5
& \textbf{83.3} & 1357.0
& 88.4 & 142.3
& \textbf{73.6} & 1780.9 \\
\bottomrule
\end{tabular}
}
\caption{Main results comparing accuracy and output token usage across three
multimodal MoE reasoning models and six benchmarks. Acc denotes answer accuracy
(\%), and Tok denotes the average number of tokens generated per sample,
including all tokens in the reasoning process and the final answer.}
\label{tab:main}
\end{table*}

Despite differences in expert count, top-$k$ configuration, and
shared-expert design, Kimi-VL, Qwen3-VL, and ERNIE-4.5-VL exhibit the same
trend, indicating that the gains are not tied to a particular MoE routing
architecture.

\paragraph{Baseline comparison.}
NoThinking, TALE, and NoWait mainly reduce computation by shortening,
bypassing, or constraining the explicit reasoning process. When
macro-averaged over the six benchmarks, NoThinking reduces output tokens by
more than 60\% for each model but lowers average accuracy by 4.0--9.4 points.
TALE and NoWait are less aggressive, yet their average accuracy also remains
below Vanilla.

Compared with RICE, TempoMoE achieves higher accuracy in all 18 settings.
Across the three models, it improves macro-averaged accuracy by
2.5--3.9 points with comparable or lower macro-averaged output token counts.
TempoMoE also improves both metrics over RICE in 10 of the 18 settings.
For example, on MathVista with Qwen3-VL, TempoMoE improves accuracy from
RICE's 76.5\% to 82.9\% with similar output lengths. Compared with Vanilla,
TempoMoE improves accuracy from 81.9\% to 82.9\% while reducing the average
output from 1350.7 to 1172.9 tokens, demonstrating a favorable trade-off
between accuracy and generation cost.

\paragraph{Task analysis.}
The size of the efficiency gain varies by task. On OCRBench, which emphasizes
visual perception and direct recognition, TempoMoE improves accuracy over
Vanilla by 1.4--2.6 points and reduces output tokens by 21.4--35.4\% across
the three models. It also achieves higher accuracy than RICE on this
benchmark for all three models.

On the more reasoning-intensive MathVista and MMMU, TempoMoE also improves
both metrics over Vanilla. Across the six corresponding settings, accuracy
increases by 0.5--1.4 points while output tokens decrease by 7.9--19.5\%.
Since the original outputs on these benchmarks are considerably longer,
even moderate relative reductions yield meaningful absolute savings.
On RealWorldQA, ChartQA, and MMStar, token reductions vary by model,
while accuracy improves consistently over Vanilla.

\begin{figure*}[!t]
    \centering
    \includegraphics[width=\textwidth]{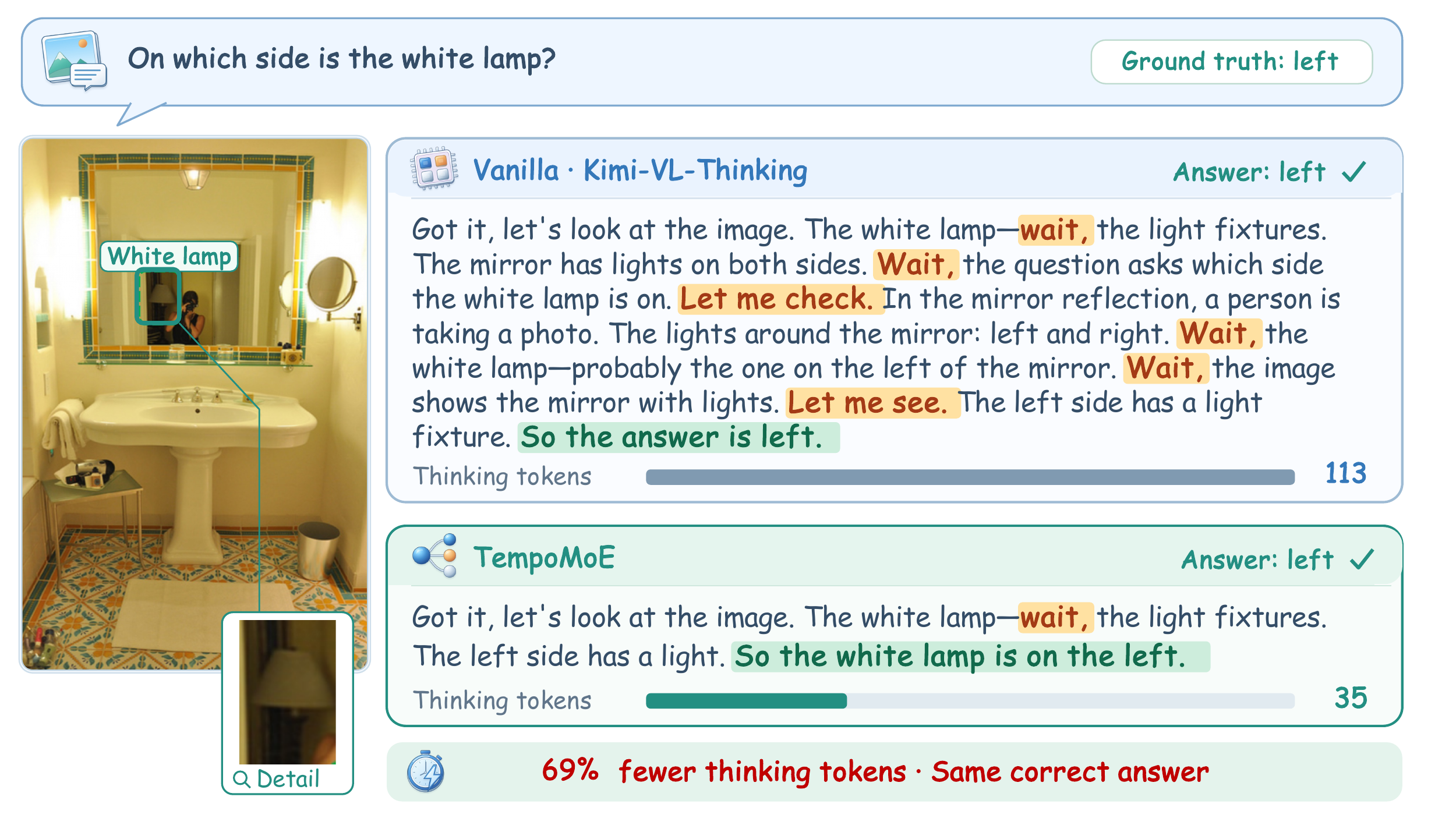}
    \caption{Visual localization case with Kimi-VL-A3B-Thinking-2506.
TempoMoE gives the same correct answer with 35 reasoning tokens instead
of 113 (69\% fewer).}
    \label{fig:case}
\end{figure*}

Figure~\ref{fig:case} illustrates this effect on a visual localization
question. Vanilla repeatedly checks the white lamp's location and describes
surrounding details before reaching the correct answer. TempoMoE gives the
same correct answer, ``left,'' while reducing thinking tokens from 113 to 35,
a reduction of approximately 69\%. This example illustrates how TempoMoE
reduces repeated checking while preserving answer correctness.

\subsection{Ablation Study}
\label{sec:exp_ablation}

We conduct ablation studies to evaluate the contributions of expert
identification, difficulty-adaptive modulation direction, and
top-$k$-preserving soft reweighting. The results are reported in
Table~\ref{tab:ablation}.

\begin{table}[!htbp]
\centering
{\small
\setlength{\tabcolsep}{2pt}
\begin{tabular}{lcccccc}
\toprule
\multirow{2}{*}{Method}
& \multicolumn{3}{c}{Accuracy (\%) $\uparrow$}
& \multicolumn{3}{c}{Output Tokens $\downarrow$} \\
\cmidrule(lr){2-4}
\cmidrule(lr){5-7}
& Overall & Easy & Hard & Overall & Easy & Hard \\
\midrule
Vanilla
& 75.6 & 80.2 & 64.5 & 625.5 & 282.6 & 1581.2 \\
Random Experts
& 72.1 & 77.8 & 59.0 & 620.2 & 280.3 & 1510.2 \\
Fixed-Fast
& 73.7 & 81.7 & 56.2 & 556.0 & 232.1 & 1408.8 \\
Fixed-Slow
& 74.8 & 79.1 & 64.9 & 679.4 & 341.6 & 1464.6 \\
Hard Mask
& 70.2 & 76.4 & 55.7 & 662.8 & 304.9 & 1600.0 \\
\midrule
\textbf{TempoMoE}
& 76.8 & 81.7 & 64.9 & 556.1 & 232.1 & 1464.6 \\
\bottomrule
\end{tabular}
}
\caption{Ablation results on Kimi-VL-A3B-Thinking-2506, aggregated across six
benchmarks. Easy and Hard use the difficulty split defined in
Section~\ref{sec:exp_setup}.}
\label{tab:ablation}
\end{table}

Random Experts replaces the identified fast- and slow-thinking experts with
the same number of randomly selected experts from the same layers, matched by
routing usage, while retaining the same modulation rule.
Fixed-Fast and Fixed-Slow retain the sample-dependent modulation magnitude
but force the modulation direction toward fast or slow thinking,
respectively, for all inputs. Hard Mask uses the same identified experts
and difficulty-conditioned direction but masks the target experts before
top-$k$ selection instead of softly reweighting them after selection.

\paragraph{Expert identification.}
Random Experts leaves the average output-token count nearly unchanged from
Vanilla but substantially lowers overall accuracy. Compared with Random
Experts, TempoMoE improves overall accuracy by 4.7 points and reduces the
average output by 64.1 tokens. This shows that the gains of TempoMoE depend
on the identified fast- and slow-thinking experts rather than generic
routing perturbations.

\paragraph{Adaptive direction.}
Fixed-Fast matches TempoMoE on easy samples but lowers accuracy on hard
samples by 8.7 points, whereas Fixed-Slow matches TempoMoE on hard samples
but generates 109.5 more tokens on average on easy samples. Since both
variants retain the sample-dependent modulation magnitude, these results
isolate the effect of adapting the modulation direction. TempoMoE selects
the direction according to sample difficulty, achieving the highest
overall accuracy while reducing output tokens. A single global direction
therefore cannot accommodate inputs with different reasoning demands.

\paragraph{Soft reweighting.}
Hard Mask underperforms Vanilla in accuracy and produces longer outputs,
indicating that changing expert selection before top-$k$ routing is
overly disruptive. TempoMoE preserves the original expert set and
redistributes routing weights, providing less disruptive control.

\section{Conclusion}

We present TempoMoE, a training-free, two-stage framework for
identifying and modulating fast- and slow-thinking experts in multimodal
MoE reasoning models. Through controlled reasoning-tempo comparisons, robustness screening,
and role-consistent interventions, we uncovered expert routing patterns
associated with different reasoning tempos and demonstrated causal effects on output length, consistent with their
tempo-associated roles. Building on these findings, TempoMoE performs
difficulty-adaptive reweighting within the original top-$k$ expert set, without
updating model parameters or altering expert selection. Experiments across
three models with distinct MoE routing designs and six benchmarks show that
TempoMoE consistently improves the accuracy--efficiency trade-off, with accuracy gains of up to 2.6 points and output-token reductions of up
to 35.4\%.
More broadly, our results show that expert-level routing mechanisms can serve
not only as explanatory tools, but also as actionable control signals for
efficient and adaptive multimodal reasoning.

\section*{Limitations}

TempoMoE requires access to internal expert routing and model-specific offline expert identification and difficulty calibration, and therefore primarily applies to multimodal MoE models with accessible internal routing. We evaluate TempoMoE on three models and six benchmarks; its performance across a broader range of model architectures and task distributions remains to be explored. Expert profiles and calibration mappings may depend on the data and prompt formulations, and recalibration may be needed under substantial distribution shifts. We use output token counts as a proxy for generation cost, while actual runtime efficiency also depends on the inference implementation and hardware environment.

\bibliography{refs}

@inproceedings{shazeer2017sparsely,
  title        = {Outrageously Large Neural Networks: The Sparsely-Gated Mixture-of-Experts Layer},
  author       = {Shazeer, Noam and Mirhoseini, Azalia and Maziarz, Krzysztof and Davis, Andy and Le, Quoc V. and Hinton, Geoffrey E. and Dean, Jeff},
  booktitle    = {International Conference on Learning Representations},
  year         = {2017},
  doi          = {10.48550/arXiv.1701.06538}
}

@article{fedus2022switch,
  title        = {Switch Transformers: Scaling to Trillion Parameter Models with Simple and Efficient Sparsity},
  author       = {Fedus, William and Zoph, Barret and Shazeer, Noam},
  journal      = {Journal of Machine Learning Research},
  volume       = {23},
  number       = {120},
  pages        = {1--39},
  year         = {2022},
  url          = {https://www.jmlr.org/papers/v23/21-0998.html}
}

@article{lin2024moellava,
  title        = {{MoE-LLaVA}: Mixture of Experts for Large Vision-Language Models},
  author       = {Lin, Bin and Tang, Zhenyu and Ye, Yang and Huang, Jinfa and Zhang, Junwu and Pang, Yatian and Jin, Peng and Ning, Munan and Luo, Jiebo and Yuan, Li},
  journal      = {IEEE Transactions on Multimedia},
  volume       = {28},
  pages        = {4408--4419},
  year         = {2026},
  doi          = {10.1109/TMM.2026.3654458}
}

@article{kimiteam2025kimivl,
  title        = {{Kimi-VL} Technical Report},
  author       = {{Kimi Team}},
  journal      = {arXiv preprint arXiv:2504.07491},
  year         = {2025},
  doi          = {10.48550/arXiv.2504.07491}
}

@article{bai2025qwen3vl,
  title        = {{Qwen3-VL} Technical Report},
  author       = {Bai, Shuai and Cai, Yuxuan and Chen, Ruizhe and Chen, Keqin and Chen, Xionghui and Cheng, Zesen and Deng, Lianghao and Ding, Wei and Gao, Chang and Ge, Chunjiang and Ge, Wenbin and Guo, Zhifang and Huang, Qidong and Huang, Jie and Huang, Fei and Hui, Binyuan and Jiang, Shutong and Li, Zhaohai and Li, Mingsheng and Li, Mei and Li, Kaixin and Lin, Zicheng and Lin, Junyang and Liu, Xuejing and Liu, Jiawei and Liu, Chenglong and Liu, Yang and Liu, Dayiheng and Liu, Shixuan and Lu, Dunjie and Luo, Ruilin and Lv, Chenxu and Men, Rui and Meng, Lingchen and Ren, Xuancheng and Ren, Xingzhang and Song, Sibo and Sun, Yuchong and Tang, Jun and Tu, Jianhong and Wan, Jianqiang and Wang, Peng and Wang, Pengfei and Wang, Qiuyue and Wang, Yuxuan and Xie, Tianbao and Xu, Yiheng and Xu, Haiyang and Xu, Jin and Yang, Zhibo and Yang, Mingkun and Yang, Jianxin and Yang, An and Yu, Bowen and Zhang, Fei and Zhang, Hang and Zhang, Xi and Zheng, Bo and Zhong, Humen and Zhou, Jingren and Zhou, Fan and Zhou, Jing and Zhu, Yuanzhi and Zhu, Ke},
  journal      = {arXiv preprint arXiv:2511.21631},
  year         = {2025},
  doi          = {10.48550/arXiv.2511.21631}
}

@inproceedings{xiao2025fastslow,
  title        = {Fast-Slow Thinking {GRPO} for Large Vision-Language Model Reasoning},
  author       = {Xiao, Wenyi and Gan, Leilei},
  booktitle    = {Advances in Neural Information Processing Systems},
  volume       = {38},
  pages        = {171601--171631},
  year         = {2025},
  publisher    = {Curran Associates, Inc.},
  doi          = {10.52202/085713-5712}
}

@inproceedings{diao2026gpro,
  title        = {Addressing Overthinking in Large Vision-Language Models via Gated Perception-Reasoning Optimization},
  author       = {Diao, Xingjian and Liu, Zheyuan and Zhang, Chunhui and Wu, Weiyi and Kong, Keyi and Shi, Lin and Ding, Kaize and Vosoughi, Soroush and Gui, Jiang},
  booktitle    = {Findings of the Association for Computational Linguistics: ACL 2026},
  pages        = {4393--4410},
  year         = {2026},
  publisher    = {Association for Computational Linguistics},
  doi          = {10.18653/v1/2026.findings-acl.215}
}

@inproceedings{huang2024harder,
  title        = {Harder Task Needs More Experts: Dynamic Routing in {MoE} Models},
  author       = {Huang, Quzhe and An, Zhenwei and Zhuang, Nan and Tao, Mingxu and Zhang, Chen and Jin, Yang and Xu, Kun and Xu, Kun and Chen, Liwei and Huang, Songfang and Feng, Yansong},
  booktitle    = {Proceedings of the 62nd Annual Meeting of the Association for Computational Linguistics (Volume 1: Long Papers)},
  pages        = {12883--12895},
  year         = {2024},
  publisher    = {Association for Computational Linguistics},
  doi          = {10.18653/v1/2024.acl-long.696}
}

@inproceedings{fayyaz2025steering,
  title        = {Steering {MoE} {LLM}s via Expert (De)Activation},
  author       = {Fayyaz, Mohsen and Modarressi, Ali and Deilamsalehy, Hanieh and Dernoncourt, Franck and Rossi, Ryan and Bui, Trung and Sch{\"u}tze, Hinrich and Peng, Nanyun},
  booktitle    = {International Conference on Learning Representations},
  year         = {2026},
  doi          = {10.48550/arXiv.2509.09660}
}

@article{hu2026activated,
  title        = {What Gets Activated: Uncovering Domain and Driver Experts in {MoE} Language Models},
  author       = {Hu, Guimin and Li, Meng and Peng, Qiwei and Hu, Lijie and Xu, Boyan and Cai, Ruichu},
  journal      = {arXiv preprint arXiv:2601.10159},
  year         = {2026},
  doi          = {10.48550/arXiv.2601.10159}
}

@inproceedings{li2025r2t2,
  title        = {{R2-T2}: Re-Routing in Test-Time for Multimodal Mixture-of-Experts},
  author       = {Li, Zhongyang and Li, Ziyue and Zhou, Tianyi},
  booktitle    = {Proceedings of the 42nd International Conference on Machine Learning},
  volume       = {267},
  series       = {Proceedings of Machine Learning Research},
  pages        = {35292--35316},
  year         = {2025},
  publisher    = {PMLR},
  url          = {https://proceedings.mlr.press/v267/li25bc.html}
}

@inproceedings{chen2024overthinking,
  title        = {Do {NOT} Think That Much for 2+3=? On the Overthinking of Long Reasoning Models},
  author       = {Chen, Xingyu and Xu, Jiahao and Liang, Tian and He, Zhiwei and Pang, Jianhui and Yu, Dian and Song, Linfeng and Liu, Qiuzhi and Zhou, Mengfei and Zhang, Zhuosheng and Wang, Rui and Tu, Zhaopeng and Mi, Haitao and Yu, Dong},
  booktitle    = {Proceedings of the 42nd International Conference on Machine Learning},
  volume       = {267},
  series       = {Proceedings of Machine Learning Research},
  pages        = {9487--9499},
  year         = {2025},
  publisher    = {PMLR},
  url          = {https://proceedings.mlr.press/v267/chen25bx.html}
}

@article{sui2025stopoverthinking,
  title        = {Stop Overthinking: A Survey on Efficient Reasoning for Large Language Models},
  author       = {Sui, Yang and Chuang, Yu-Neng and Wang, Guanchu and Zhang, Jiamu and Zhang, Tianyi and Yuan, Jiayi and Liu, Hongyi and Wen, Andrew and Zhong, Shaochen and Zou, Na and Chen, Hanjie and Hu, Xia},
  journal      = {Transactions on Machine Learning Research},
  year         = {2025},
  doi          = {10.48550/arXiv.2503.16419}
}

@inproceedings{huang2026cgrs,
  title        = {Efficient Reasoning for Large Reasoning Language Models via Certainty-Guided Reflection Suppression},
  author       = {Huang, Jiameng and Lin, Baijiong and Feng, Guhao and Chen, Jierun and He, Di and Hou, Lu},
  booktitle    = {Proceedings of the AAAI Conference on Artificial Intelligence},
  volume       = {40},
  number       = {37},
  pages        = {31176--31184},
  year         = {2026},
  doi          = {10.1609/aaai.v40i37.40379}
}

@inproceedings{yang2025deer,
  title        = {Dynamic Early Exit in Reasoning Models},
  author       = {Yang, Chenxu and Si, Qingyi and Duan, Yongjie and Zhu, Zheliang and Zhu, Chenyu and Li, Qiaowei and Chen, Minghui and Lin, Zheng and Wang, Weiping},
  booktitle    = {International Conference on Learning Representations},
  year         = {2026},
  doi          = {10.48550/arXiv.2504.15895}
}

@inproceedings{han2025tale,
  title        = {Token-Budget-Aware {LLM} Reasoning},
  author       = {Han, Tingxu and Wang, Zhenting and Fang, Chunrong and Zhao, Shiyu and Ma, Shiqing and Chen, Zhenyu},
  booktitle    = {Findings of the Association for Computational Linguistics: ACL 2025},
  pages        = {24842--24855},
  year         = {2025},
  publisher    = {Association for Computational Linguistics},
  doi          = {10.18653/v1/2025.findings-acl.1274}
}

@inproceedings{wang2024esft,
  title        = {Let the Expert Stick to His Last: Expert-Specialized Fine-Tuning for Sparse Architectural Large Language Models},
  author       = {Wang, Zihan and Chen, Deli and Dai, Damai and Xu, Runxin and Li, Zhuoshu and Wu, Yu},
  booktitle    = {Proceedings of the 2024 Conference on Empirical Methods in Natural Language Processing},
  pages        = {784--801},
  year         = {2024},
  publisher    = {Association for Computational Linguistics},
  doi          = {10.18653/v1/2024.emnlp-main.46}
}

@inproceedings{arora2025efficient,
  title        = {Training Language Models to Reason Efficiently},
  author       = {Arora, Daman and Zanette, Andrea},
  booktitle    = {Advances in Neural Information Processing Systems},
  volume       = {38},
  pages        = {60770--60808},
  year         = {2025},
  publisher    = {Curran Associates, Inc.},
  doi          = {10.52202/085713-2032}
}

@inproceedings{aggarwal2025l1,
  title        = {{L1}: Controlling How Long a Reasoning Model Thinks with Reinforcement Learning},
  author       = {Aggarwal, Pranjal and Welleck, Sean},
  booktitle    = {Conference on Language Modeling},
  year         = {2025},
  doi          = {10.48550/arXiv.2503.04697}
}

@inproceedings{fu2024dynasor,
  title        = {Efficiently Scaling {LLM} Reasoning Programs with {Certaindex}},
  author       = {Fu, Yichao and Chen, Junda and Zhu, Siqi and Fu, Zheyu and Dai, Zhongdongming and Zhuang, Yonghao and Ma, Yian and Qiao, Aurick and Rosing, Tajana and Stoica, Ion and Zhang, Hao},
  booktitle    = {Advances in Neural Information Processing Systems},
  volume       = {38},
  pages        = {141987--142025},
  year         = {2025},
  publisher    = {Curran Associates, Inc.},
  doi          = {10.52202/085713-4751}
}

@article{lin2025dualmindvlm,
  title        = {Learning to Think Fast and Slow for Visual Language Models},
  author       = {Lin, Chenyu and Chi, Cheng and Wu, Jinlin and Li, Sharon and Zhou, Kaiyang},
  journal      = {arXiv preprint arXiv:2511.16670},
  year         = {2025},
  doi          = {10.48550/arXiv.2511.16670}
}

@article{lan2025metishome,
  title        = {{Metis-HOME}: Hybrid Optimized Mixture-of-Experts for Multimodal Reasoning},
  author       = {Lan, Xiaohan and Liu, Fanfan and Qiu, Haibo and Yang, Siqi and Ruan, Delian and Shi, Peng and Ma, Lin},
  journal      = {arXiv preprint arXiv:2510.20519},
  year         = {2025},
  doi          = {10.48550/arXiv.2510.20519}
}

@article{ma2025nothinking,
  title        = {Reasoning Models Can Be Effective Without Thinking},
  author       = {Ma, Wenjie and He, Jingxuan and Snell, Charlie and Griggs, Tyler and Min, Sewon and Zaharia, Matei},
  journal      = {arXiv preprint arXiv:2504.09858},
  year         = {2025},
  doi          = {10.48550/arXiv.2504.09858}
}

@article{xu2025chainofdraft,
  title        = {Chain of Draft: Thinking Faster by Writing Less},
  author       = {Xu, Silei and Xie, Wenhao and Zhao, Lingxiao and He, Pengcheng},
  journal      = {arXiv preprint arXiv:2502.18600},
  year         = {2025},
  doi          = {10.48550/arXiv.2502.18600}
}

@inproceedings{wang2025nowait,
  title        = {Wait, We Don{'}t Need to ``Wait''! Removing Thinking Tokens Improves Reasoning Efficiency},
  author       = {Wang, Chenlong and Feng, Yuanning and Chen, Dongping and Chu, Zhaoyang and Krishna, Ranjay and Zhou, Tianyi},
  booktitle    = {Findings of the Association for Computational Linguistics: EMNLP 2025},
  pages        = {7459--7482},
  year         = {2025},
  publisher    = {Association for Computational Linguistics},
  doi          = {10.18653/v1/2025.findings-emnlp.394}
}

@inproceedings{muennighoff2025s1,
  title        = {{s1}: Simple test-time scaling},
  author       = {Muennighoff, Niklas and Yang, Zitong and Shi, Weijia and Li, Xiang Lisa and Fei-Fei, Li and Hajishirzi, Hannaneh and Zettlemoyer, Luke and Liang, Percy and Cand{\`e}s, Emmanuel and Hashimoto, Tatsunori},
  booktitle    = {Proceedings of the 2025 Conference on Empirical Methods in Natural Language Processing},
  pages        = {20275--20321},
  year         = {2025},
  publisher    = {Association for Computational Linguistics},
  doi          = {10.18653/v1/2025.emnlp-main.1025}
}

@article{lou2025adacot,
  title        = {{AdaCoT}: Pareto-Optimal Adaptive Chain-of-Thought Triggering via Reinforcement Learning},
  author       = {Lou, Chenwei and Sun, Zewei and Liang, Xinnian and Qu, Meng and Shen, Wei and Wang, Wenqi and Li, Yuntao and Yang, Qingping and Wu, Shuangzhi},
  journal      = {arXiv preprint arXiv:2505.11896},
  year         = {2025},
  doi          = {10.48550/arXiv.2505.11896}
}

@inproceedings{wu2025arm,
  title        = {{ARM}: Adaptive Reasoning Model},
  author       = {Wu, Siye and Xie, Jian and Zhang, Yikai and Chen, Aili and Zhang, Kai and Su, Yu and Xiao, Yanghua},
  booktitle    = {Advances in Neural Information Processing Systems},
  volume       = {38},
  pages        = {50680--50710},
  year         = {2025},
  publisher    = {Curran Associates, Inc.},
  doi          = {10.52202/085713-1691}
}

@inproceedings{lo2025closer,
  title        = {A Closer Look into Mixture-of-Experts in Large Language Models},
  author       = {Lo, Ka Man and Huang, Zeyu and Qiu, Zihan and Wang, Zili and Fu, Jie},
  booktitle    = {Findings of the Association for Computational Linguistics: NAACL 2025},
  pages        = {4427--4447},
  year         = {2025},
  publisher    = {Association for Computational Linguistics},
  doi          = {10.18653/v1/2025.findings-naacl.251}
}

@inproceedings{bandarkar2025multilingual,
  title        = {Multilingual Routing in Mixture-of-Experts},
  author       = {Bandarkar, Lucas and Yang, Chenyuan and Fayyaz, Mohsen and Hu, Junlin and Peng, Nanyun},
  booktitle    = {International Conference on Learning Representations},
  year         = {2026},
  doi          = {10.48550/arXiv.2510.04694}
}

@article{lai2025safex,
  title        = {{SAFEx}: Analyzing Vulnerabilities of {MoE}-Based {LLM}s via Stable Safety-Critical Expert Identification},
  author       = {Lai, Zhenglin and Liao, Mengyao and Wu, Bingzhe and Xu, Dong and Zhao, Zebin and Yuan, Zhihang and Fan, Chao and Li, Jianqiang},
  journal      = {arXiv preprint arXiv:2506.17368},
  year         = {2025},
  doi          = {10.48550/arXiv.2506.17368}
}

@article{xu2026seeing,
  title        = {Seeing but Not Thinking: Routing Distraction in Multimodal Mixture-of-Experts},
  author       = {Xu, Haolei and Hong, Haiwen and Li, Hongxing and Zhou, Rui and Zhang, Yang and Huang, Longtao and Xue, Hui and Shen, Yongliang and Lu, Weiming and Zhuang, Yueting},
  journal      = {arXiv preprint arXiv:2604.08541},
  year         = {2026},
  doi          = {10.48550/arXiv.2604.08541}
}

@inproceedings{wei2022chain,
  title        = {Chain-of-Thought Prompting Elicits Reasoning in Large Language Models},
  author       = {Wei, Jason and Wang, Xuezhi and Schuurmans, Dale and Bosma, Maarten and Ichter, Brian and Xia, Fei and Chi, Ed H. and Le, Quoc V. and Zhou, Denny},
  booktitle    = {Advances in Neural Information Processing Systems},
  volume       = {35},
  pages        = {24824--24837},
  year         = {2022},
  publisher    = {Curran Associates, Inc.},
  doi          = {10.52202/068431-1800}
}

@techreport{ernie2025technicalreport,
  title        = {{ERNIE} 4.5 Technical Report},
  author       = {{Baidu ERNIE Team}},
  institution  = {Baidu},
  type         = {Technical report},
  year         = {2025},
  url          = {https://ernie.baidu.com/blog/publication/ERNIE_Technical_Report.pdf}
}

@inproceedings{chen2024mmstar,
  title        = {Are We on the Right Way for Evaluating Large Vision-Language Models?},
  author       = {Chen, Lin and Li, Jinsong and Dong, Xiaoyi and Zhang, Pan and Zang, Yuhang and Chen, Zehui and Duan, Haodong and Wang, Jiaqi and Qiao, Yu and Lin, Dahua and Zhao, Feng},
  booktitle    = {Advances in Neural Information Processing Systems},
  volume       = {37},
  pages        = {27056--27087},
  year         = {2024},
  publisher    = {Curran Associates, Inc.},
  doi          = {10.52202/079017-0850}
}

@article{liu2023ocrbench,
  title        = {{OCRBench}: On the Hidden Mystery of {OCR} in Large Multimodal Models},
  author       = {Liu, Yuliang and Li, Zhang and Huang, Mingxin and Yang, Biao and Yu, Wenwen and Li, Chunyuan and Yin, Xucheng and Liu, Cheng-Lin and Jin, Lianwen and Bai, Xiang},
  journal      = {Science China Information Sciences},
  volume       = {67},
  pages        = {220102},
  year         = {2024},
  doi          = {10.1007/s11432-024-4235-6}
}

@misc{xai2024realworldqa,
  title        = {{Grok}-1.5 Vision Preview},
  author       = {{xAI}},
  howpublished = {{xAI} blog},
  year         = {2024},
  note         = {Introduces the {RealWorldQA} benchmark},
  url          = {https://x.ai/news/grok-1.5v}
}

@inproceedings{lu2024mathvista,
  title        = {{MathVista}: Evaluating Mathematical Reasoning of Foundation Models in Visual Contexts},
  author       = {Lu, Pan and Bansal, Hritik and Xia, Tony and Liu, Jiacheng and Li, Chunyuan and Hajishirzi, Hannaneh and Cheng, Hao and Chang, Kai-Wei and Galley, Michel and Gao, Jianfeng},
  booktitle    = {International Conference on Learning Representations},
  year         = {2024},
  doi          = {10.48550/arXiv.2310.02255}
}

@inproceedings{masry2022chartqa,
  title        = {{ChartQA}: A Benchmark for Question Answering about Charts with Visual and Logical Reasoning},
  author       = {Masry, Ahmed and Long, Do Xuan and Tan, Jia Qing and Joty, Shafiq and Hoque, Enamul},
  booktitle    = {Findings of the Association for Computational Linguistics: ACL 2022},
  pages        = {2263--2279},
  year         = {2022},
  publisher    = {Association for Computational Linguistics},
  doi          = {10.18653/v1/2022.findings-acl.177}
}

@inproceedings{yue2024mmmu,
  title        = {{MMMU}: A Massive Multi-discipline Multimodal Understanding and Reasoning Benchmark for Expert {AGI}},
  author       = {Yue, Xiang and Ni, Yuansheng and Zhang, Kai and Zheng, Tianyu and Liu, Ruoqi and Zhang, Ge and Stevens, Samuel and Jiang, Dongfu and Ren, Weiming and Sun, Yuxuan and Wei, Cong and Yu, Botao and Yuan, Ruibin and Sun, Renliang and Yin, Ming and Zheng, Boyuan and Yang, Zhenzhu and Liu, Yibo and Huang, Wenhao and Sun, Huan and Su, Yu and Chen, Wenhu},
  booktitle    = {Proceedings of the IEEE/CVF Conference on Computer Vision and Pattern Recognition},
  pages        = {9556--9567},
  year         = {2024},
  doi          = {10.1109/CVPR52733.2024.00913}
}

@article{li2025unimoe,
  title        = {{Uni-MoE}: Scaling Unified Multimodal {LLM}s with Mixture of Experts},
  author       = {Li, Yunxin and Jiang, Shenyuan and Hu, Baotian and Wang, Longyue and Zhong, Wanqi and Luo, Wenhan and Ma, Lin and Zhang, Min},
  journal      = {IEEE Transactions on Pattern Analysis and Machine Intelligence},
  volume       = {47},
  number       = {5},
  pages        = {3424--3439},
  year         = {2025},
  doi          = {10.1109/TPAMI.2025.3532688}
}

@article{wu2024deepseekvl2,
  title        = {{DeepSeek-VL2}: Mixture-of-Experts Vision-Language Models for Advanced Multimodal Understanding},
  author       = {Wu, Zhiyu and Chen, Xiaokang and Pan, Zizheng and Liu, Xingchao and Liu, Wen and Dai, Damai and Gao, Huazuo and Ma, Yiyang and Wu, Chengyue and Wang, Bingxuan and Xie, Zhenda and Wu, Yu and Hu, Kai and Wang, Jiawei and Sun, Yaofeng and Li, Yukun and Piao, Yishi and Guan, Kang and Liu, Aixin and Xie, Xin and You, Yuxiang and Dong, Kai and Yu, Xingkai and Zhang, Haowei and Zhao, Liang and Wang, Yisong and Ruan, Chong},
  journal      = {arXiv preprint arXiv:2412.10302},
  year         = {2024},
  doi          = {10.48550/arXiv.2412.10302}
}

@inproceedings{huang2025modes,
  title        = {{MoDES}: Accelerating Mixture-of-Experts Multimodal Large Language Models via Dynamic Expert Skipping},
  author       = {Huang, Yushi and Wang, Zining and Yuan, Zhihang and Ding, Yifu and Gong, Ruihao and Guo, Jinyang and Liu, Xianglong and Zhang, Jun},
  booktitle    = {Proceedings of the IEEE/CVF Conference on Computer Vision and Pattern Recognition},
  pages        = {30205--30215},
  year         = {2026},
  url          = {https://openaccess.thecvf.com/content/CVPR2026/html/Huang_MoDES_Accelerating_Mixture-of-Experts_Multimodal_Large_Language_Models_via_Dynamic_Expert_CVPR_2026_paper.html}
}

@inproceedings{olson2025semantic,
  title        = {Probing Semantic Routing in Large Mixture-of-Expert Models},
  author       = {Olson, Matthew Lyle and Ratzlaff, Neale and Hinck, Musashi and Luo, Man and Yu, Sungduk and Xue, Chendi and Lal, Vasudev},
  booktitle    = {Findings of the Association for Computational Linguistics: EMNLP 2025},
  pages        = {18263--18278},
  year         = {2025},
  publisher    = {Association for Computational Linguistics},
  doi          = {10.18653/v1/2025.findings-emnlp.991}
}

@article{li2025selfbudgeter,
  title        = {{SelfBudgeter}: Adaptive Token Allocation for Efficient {LLM} Reasoning},
  author       = {Li, Zheng and Dong, Qingxiu and Ma, Jingyuan and Zhang, Di and Jia, Kai and Sui, Zhifang},
  journal      = {arXiv preprint arXiv:2505.11274},
  year         = {2025},
  doi          = {10.48550/arXiv.2505.11274}
}

@inproceedings{wang2025rice,
  title        = {Two Experts Are All You Need for Steering Thinking: Reinforcing Cognitive Effort in {MoE} Reasoning Models Without Additional Training},
  author       = {Wang, Mengru and Chen, Xingyu and Wang, Yue and He, Zhiwei and Xu, Jiahao and Liang, Tian and Liu, Qiuzhi and Yao, Yunzhi and Wang, Wenxuan and Ma, Ruotian and Mi, Haitao and Zhang, Ningyu and Tu, Zhaopeng and Li, Xiaolong and Yu, Dong},
  booktitle    = {Advances in Neural Information Processing Systems},
  volume       = {38},
  pages        = {140699--140724},
  year         = {2025},
  publisher    = {Curran Associates, Inc.},
  doi          = {10.52202/085713-4701}
}

\appendix

\end{document}